\documentclass[11pt]{article}

\usepackage[final]{acl}

 \usepackage{microtype}

\usepackage[english,bidi=default]{babel} 
\babelfont{rm}{TeXGyreTermesX} 
\babelfont{tt}[Extension=.otf, UprightFont=*-Regular, BoldFont=*-Bold]{Inconsolatazi4}
\babelprovide[import]{hindi}
\babelfont[*devanagari]{rm}{Lohit Devanagari} 
\babelprovide[import]{arabic}
\babelfont[*arabic]{rm}[Extension=.ttf, UprightFont=*-Regular, BoldFont=*-Bold,
  ItalicFont=*-Italic, BoldItalicFont=*-BoldItalic]{Amiri}
\babelprovide[import]{hebrew}
\babelfont[*hebrew]{rm}[Extension=.ttf, UprightFont=*-Medium, BoldFont=*-Bold,
  ItalicFont=*-MediumOblique, BoldItalicFont=*-BoldOblique]{FrankRuehlCLM}
\babelprovide[import]{chinese}
\babelfont[*cjk]{rm}[Path=fonts/, Extension=.otf,
  UprightFont=NotoSerifCJK-Subset]{NotoSerifCJKCombined}
\babelprovide[import]{korean}
\babelfont[*hangul]{rm}[Path=fonts/, Extension=.otf,
  UprightFont=NotoSerifCJK-Subset]{NotoSerifCJKCombined}

\newfontfamily\cjkfont[Path=fonts/, Extension=.otf,
  UprightFont=NotoSerifCJK-Subset]{NotoSerifCJKCombined}

\newfontfamily\hangulfont[Path=fonts/, Extension=.otf,
  UprightFont=NotoSerifCJK-Subset]{NotoSerifCJKCombined}

\usepackage{latexsym}
\usepackage{amsmath,amsfonts,amssymb}
\usepackage{bbm}
\usepackage{listings}
\usepackage{xcolor}
\usepackage{fvextra}
\usepackage{subcaption}
\usepackage{booktabs}
\usepackage{xspace}

\usepackage{graphicx}

\newcommand{\QwenLargeInstruct}{\text{Qwen3-30B-A3B-Instruct-2507}\xspace}
\newcommand{\QwenLargeThink}{\text{Qwen3-30B-A3B-Thinking-2507}\xspace}
\newcommand{\QwenSmallInstruct}{\text{Qwen3-4B-Instruct-2507}\xspace}
\newcommand{\QwenSmallThink}{\text{Qwen3-4B-Thinking-2507}\xspace}
\newcommand{\Llama}{\text{Llama-3.2-3B-Instruct}\xspace}
\newcommand{\Gemma}{\text{gemma-3-27b-it}\xspace}
\newcommand{\QwenOld}{\text{Qwen2.5-32B-Instruct}\xspace}
\newcommand{\DeepseekQwen}{\text{DeepSeek-R1-Distill-Qwen-32B}\xspace}
\newcommand{\Aya}{\text{aya-expanse-32b}\xspace}
\newcommand{\Tower}{\text{Tower-Plus-72B}\xspace}
\newcommand{\TFiveGemma}{\text{t5gemma-xl-xl-prefixlm-it}\xspace}
\newcommand{\DeepseekChat}{\text{DeepSeek-V3.2-Exp-671B-chat}\xspace}

\title{TransClean: A Benchmark for Detecting and Extracting Clean Translations from Large Language Model Outputs}

\author{
Shenbin Qian \and Yves Scherrer \\
Language Technology Group, Department of Informatics \\
University of Oslo, Norway \\
\texttt{\{shenbinq, yves.scherrer\}@ifi.uio.no}
}

\begin{document}

\maketitle
\begin{abstract}
Large language models (LLMs) are increasingly used for machine translation, yet their outputs often contain additional text beyond the translation itself, such as language labels, explanations or bilingual repetitions, which we term \textit{translation noise}. Despite its prevalence, this problem lacks dedicated benchmarks and systematic study. We analyze over 790,000 translation outputs from 12 LLMs across 22 language pairs (LPs) and identify 12 recurring noise patterns, which we group into \textit{formatting}  and \textit{content} noise. Building on the observed patterns, we construct \textbf{TransClean}, a controlled benchmark of 9,900 pairs of noisy and clean translation outputs, comprising 8,800 \textit{synthetically generated} instances and 1,100 manually curated \textit{authentic} instances. 
We evaluate two extraction approaches on the Trans\-Clean benchmark: 1) a span-based extraction method leveraging translation quality estimation models for span detection, and 2) an LLM-based extraction method that prompts an LLM to isolate the translation. Our benchmark and analysis provide the first systematic framework to evaluate and improve the cleanliness of LLM translation outputs.
\end{abstract}

\section{Introduction}

Large language models (LLMs) are rapidly reshaping the landscape of machine translation (MT). LLMs can perform high-quality translation through prompting alone and increasingly match or even surpass task-specific MT systems in many scenarios \citep{ZhangBiao2023,vilar-etal-2023-prompting,kocmi-etal-2024-findings,xu2024}. The flexibility and multilingual capacity of LLMs have led to widespread adoption in both research and deployment settings, from systems developed for the Conference on Machine Translation (WMT\footnote{\url{https://www2.statmt.org/}}) shared tasks \citep{kocmi-etal-2025-findings} to large-scale commercial platforms such as Google Translate \citep{Caswell22024} and social media services like Instagram \citep{Instagram2026}. As LLMs increasingly serve as translation engines, understanding and standardizing their outputs becomes critical.

\begin{figure*}[t]
  \centering
  \includegraphics[width=\textwidth]{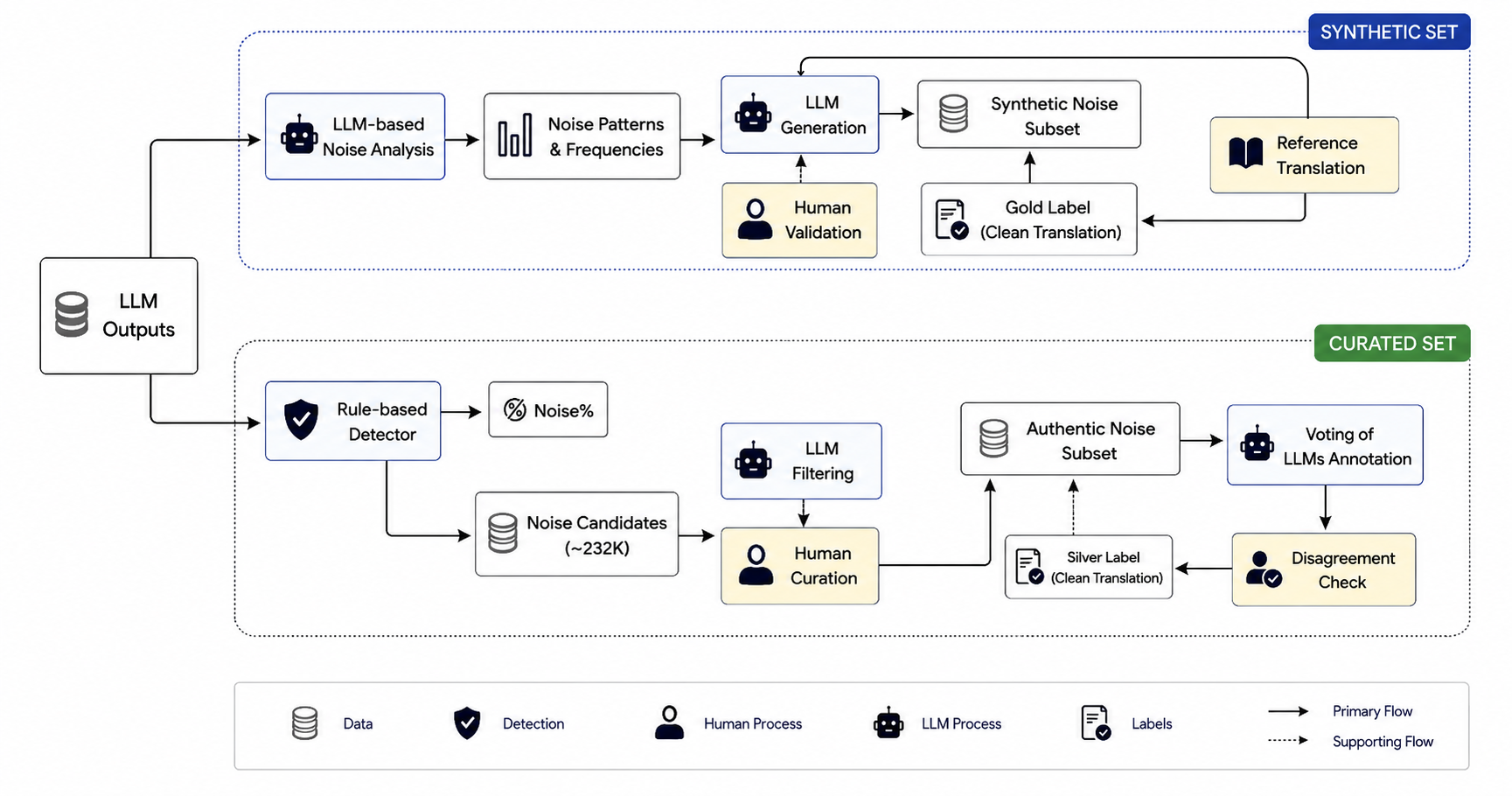}
  \caption{The process of creating \textbf{TransClean} to benchmark noise detection and clean translation extraction.}
  \label{fig.pipeline}
\end{figure*}

However, LLM translations often contain additional text beyond the translation itself. Instead of producing a single target-language translation, models may prepend language labels, append explanations, repeat the source sentence, or provide cultural commentary. While such behavior can be helpful in interactive settings, it introduces a systematic challenge for automatic evaluation and downstream integration. Standard MT evaluation pipelines assume that model outputs consist solely of the translation. Extra content can distort metric scores and introduce inconsistencies in large-scale benchmarking. We refer to this phenomenon as \textit{translation noise}: any content in an LLM output that is not part of the intended target translation. 

In preliminary experiments across multiple models and prompts, we observe that translation noise is not rare. For most LLMs we evaluate, 3\% to 99\%\footnote{We formally define and quantify the noise rate in $\S$\ref{sec:noise_rate}.} of the translation outputs contain additional explanatory or formatting text. Although carefully engineered prompts (e.g., ``Output the translation only.'') reduce this behavior, they do not fully eliminate it. The prevalence and form of noise vary substantially across models, reflecting differences in instruction-following abilities. As a result, clean translation cannot be reliably guaranteed through prompting alone.

Despite its practical importance, translation noise has not been systematically studied. Prior work has examined related issues such as instruction following in multilingual settings \citep{Lietal2024}, instruction forgetting \citep{Chen2023-cr}, and undesirable behaviors such as repetitive texts or wrong target-language outputs \citep{bawden-yvon-2023-investigating,wang-etal-2024-mitigating-language}. Existing studies have largely centered either on mitigating these issues through model modification or fine-tuning, or on evaluating instruction-following capabilities by proposing new benchmarks such as IFEval \citep{zhou2023instructionfollowingevaluationlargelanguage}, InFoBench \citep{qin-etal-2024-infobench}, and M-IFEval \citep{dussolle-etal-2025-ifeval}. However, little attention has been paid to analyzing noise patterns directly in LLM translation outputs or to developing post-processing methods that recover clean translations without altering the underlying models. This distinction is crucial in real-world deployment, where models may be proprietary, closed-source, or too costly to retrain.

In this work, we formalize the task of \textbf{clean translation extraction}: given an LLM output that may contain translation noise, extract the span corresponding to the correct target-language translation. We approach this task in three steps:

\begin{enumerate}
    \item We conduct a large-scale empirical study ($\S$\ref{sec:trans_noise}) of translation noise in LLM outputs, analyzing over 790,000 translations from 12 LLMs across 22 language pairs and identifying 12 recurring noise patterns and 2 main categories.
    \item We construct \textbf{TransClean}\footnote{\url{https://github.com/shenbinqian/TransClean}}, the first benchmark ($\S$\ref{sec:bench_build}) for clean translation extraction, comprising 8,800 synthetically noised instances and 1,100 manually curated authentic noisy examples with silver clean translations. The process of creating \textbf{TransClean} is illustrated in Figure \ref{fig.pipeline}.
    \item Using TransClean, we benchmark two approaches to \textbf{extract clean translations} ($\S$\ref{sec:trans_ext}): 1) a span-based extraction method leveraging quality estimation models for span detection, 2) an LLM-based extractor that prompts an LLM to isolate the translation. In this context, we design two evaluation metrics that enable standardized comparison.
\end{enumerate}


\begin{table*}[h]
\centering
\resizebox{0.95\linewidth}{!}{%
\begin{tabular}{lrrr|rrr|rrr}
\toprule
                                         & \multicolumn{3}{c}{Noise\% $\uparrow$}                            & \multicolumn{3}{c}{Expl\% $\uparrow$}                          & \multicolumn{3}{c}{WrongL\% $\uparrow$}                  \\
Model name                              & Prompt 0         & Prompt 1         & Prompt 2         & Prompt 0        & Prompt 1        & Prompt 2        & Prompt 0 & Prompt 1        & Prompt 2         \\ \midrule
\QwenLargeInstruct         & \textbf{4.02\%}  & 3.31\%          & 3.28\%           & \textbf{2.95\%} & 2.69\%          & 2.62\%          & \textbf{1.18\%}   & 0.64\% & 0.68\%           \\
\QwenLargeThink         & \textbf{3.30\%}  & 3.26\%          & 3.28\%          & \textbf{2.81\%} & 2.76\%          & 2.78\%          & \textbf{0.65\%}   & 0.51\% & 0.52\%           \\
\QwenSmallInstruct              & \textbf{4.65\%}  & 3.77\%          & 3.79\%          & 3.03\%          & \textbf{3.09\%} & 3.08\%          & \textbf{1.77\%}   & 0.70\% & 0.77\%           \\
\QwenSmallThink              & 3.51\%           & \textbf{3.66\%} & 3.62\%          & 2.84\%          & \textbf{2.94\%} & 2.91\%          & \textbf{0.99\%}   & 0.75\%          & 0.73\%  \\
\Llama         & 40.58\%          & \textbf{69.54\%} & 7.09\%          & 29.47\%         & \textbf{67.60\%} & 3.30\%         & \textbf{21.67\%}  & 15.87\%         & 4.03\%  \\
\Gemma                   & 97.32\%           & \textbf{99.73\%} & 3.24\%          & 97.32\%         & \textbf{99.72\%} & 2.79\%         & \textbf{31.82\%}  & \textbf{31.82\%} & 0.46\%  \\
\QwenOld                & \textbf{55.62\%} & 28.05\%          & 9.06\%          & \textbf{53.93\%} & 26.71\%         & 7.21\%         & \textbf{15.22\%}  & 8.08\%          & 4.10\%  \\
\DeepseekQwen & \textbf{26.75\%} & 15.90\%          & 4.19\%          & \textbf{22.22\%} & 14.42\%         & 2.89\%         & \textbf{11.92\%}  & 5.18\%          & 1.36\%  \\
\Aya               & 16.63\%          & \textbf{31.71\%} & 3.65\%          & 15.03\%         & \textbf{27.46\%} & 3.10\%         & 4.93\%   & \textbf{12.47\%} & 0.60\%  \\
\Tower                   & 5.15\%          & \textbf{5.26\%}  & 3.88\%          & 3.34\%          & \textbf{4.58\%} & 3.33\%          & \textbf{2.05\%}   & 0.94\%          & 0.58\%  \\
\TFiveGemma         & \textbf{46.32\%} & 27.64\%          & 17.35\%         & \textbf{33.88\%} & 12.86\%         & 4.11\%         & \textbf{25.41\%}  & 19.35\%         & 14.23\% \\
\DeepseekChat             & \textbf{60.72\%} & \multicolumn{1}{c}{/}                & 3.19\%          & \textbf{59.17\%} & \multicolumn{1}{c}{/}                & 2.86\%         & \textbf{11.65\%}  & \multicolumn{1}{c}{/}                & 0.37\%  \\
\bottomrule
\end{tabular}%
}
\caption{The noise rate (Noise\%), the rate of generating explanatory texts (Expl\%) and the rate of outputting the wrong target language (WrongL\%) for different prompts and LLMs. We did not run all prompts on DeepSeek-V3.2-Exp as we see Prompt 0 generally leads to a higher noise rate across models.}
\label{tab:clean_rate}
\end{table*}

\section{Noise in LLM Translation Outputs} \label{sec:trans_noise}

In order to assess the prevalence and types of noise present in LLM-produced translations, we generate a large sample of translations for 22 language pairs (LPs) using 12 representative LLMs and 3 prompt templates ($\S$\ref{sec:data_src}). We then systematically examine these translation outputs and analyze their noise rate and patterns ($\S$\ref{sec:noise_rate} and $\S$\ref{sec:noise_analysis}).


\subsection{Generating Noisy Translations} \label{sec:data_src}


\paragraph{Data Sources} We identify 22 LPs with varying resource levels and translation directions and randomly sample 3,000 sentence pairs per LP from four parallel corpora collections: the TED Multilingual Parallel Corpus \citep{Kulkarni2015}, the WMT20 Quality Estimation Dataset \citep{barrault-etal-2020-findings}, the SwissAdmin corpus \citep{scherrer-etal-2014-swissadmin} and the Chinese--Korean parallel corpus \citep{Park2019-vd}, resulting in a test set of 66,000 instances in total. Detailed information on LPs, dataset sizes, and their corresponding sources is provided in Table~\ref{tab:data_details} in the appendix.

\paragraph{Prompt Templates} 
We design three prompt templates (see Figure~\ref{promt_trans}) to investigate how prompting strategies influence the generation of noisy translations and to identify which prompt yields the highest number of noisy instances. Prompt 0 is adopted from \citet{ZhangBiao2023}, while Prompt 1 and Prompt 2 are newly designed templates.

\begin{figure}[h]
\centering

\begin{tabular}{|p{6.5cm}|}
\hline
\textbf{Prompt 0} \\
    
    \{src\_lang\}: \{src\_txt\} \\
    \{tgt\_lang\}: \\

\hline
    
    \textbf{Prompt 1} \\
    
    Translate the following \{src\_lang\} into \{tgt\_lang\}: \{src\_text\} \\

\hline
    
    \textbf{Prompt 2} \\
    
    Translate the following \{src\_lang\} into \{tgt\_lang\} and only output the target text: \{src\_text\} \\
\hline
\end{tabular}

  \caption{Prompt templates for translation.}
\label{promt_trans}
\end{figure}

\paragraph{Model Selection} 

Different LLMs may exhibit varying tendencies in producing noisy outputs. To capture this variability, we select 12 open-weights LLMs that span a diverse range of model sizes, architectures, post-training methods, and multilingual training coverage. The selected models include decoder-only instruction-tuned models and their reasoning variants, like \QwenSmallInstruct and \QwenSmallThink \citep{QwenTeam2025}; large frontier mixture-of-experts models such as DeepSeek-V3.2-Exp \citep{deepseekai2024deepseekv32}; smaller dense models including \Llama \citep{meta2024llamav32} and \Gemma \citep{Gemma_Team2025-ga}; recently released instruction-tuned encoder-decoder models such as \TFiveGemma \citep{Zhang2025-ya}; multilingual models including \Aya \citep{Dang2024-nk}; and \Tower \citep{Rei2025-ij}, a translation-oriented LLM fine-tuned on Qwen-2.5-72B \citep{QwenTeam2024}. Details of all our models can be found in Table~\ref{tab:model_details} in the appendix.

\paragraph{Inference} We generate the translation outputs exclusively in a zero-shot setting. The details of LLM inference used to generate the translation outputs are provided in Appendix~\ref{sec:appendix_inf}.

\begin{table*}[]
\centering
\resizebox{0.95\linewidth}{!}{%
\begin{tabular}{lcrl}
\toprule
Noise pattern       & Examples     & Frequency & Category   \\ \midrule
\texttt{explanation}         & \begin{tabular}[c]{@{}c@{}}\foreignlanguage{chinese}{这三名红军战士住在一个没有危险的小村庄里。}\textcolor{red}{\textbackslash{}n\textbackslash{}n**Explanation:**\textbackslash{}n\textbackslash{}n*   **\foreignlanguage{korean}{이 세 명의 홍군 전사는}**}\\  \textcolor{red}{- This translates to \textbackslash{}"These three Hongjun   warriors\textbackslash{}" \textbackslash{}n...}\end{tabular} & 33\%        & content    \\ 
\texttt{alternatives}        & \begin{tabular}[c]{@{}c@{}}This traffic accident caused chaos in the railway traffic entering the city.\textcolor{red}{\textbackslash{}n\textbackslash{}nAlternative translation:\textbackslash{}n}\\ \textcolor{red}{\foreignlanguage{chinese}{这次交通事故使该市进入铁路交通陷入混乱。}}\end{tabular}                                                             & 15\%        & content    \\
\texttt{off-topic}          & \begin{tabular}[c]{@{}c@{}}I am an AI assistant designed to be helpful. I can provide information, answer questions, \\ and help with tasks to the best of my abilities. How can I help you today?\end{tabular}                                                                    & 12\%        & content    \\
\texttt{verbose preamble}   & \textcolor{red}{Here is the translation of} \textbackslash{}``Vulnerable Dems air impeachment concerns to Pelosi'' \textcolor{red}{from English to Chinese:} ...                                                                                                                                                       & 9\%         & formatting \\
\texttt{bilingual output}   & \textcolor{red}{Arabic: \foreignlanguage{arabic}{أنا أحب الحيوانات} \textbackslash{}nTranslation:}\foreignlanguage{hebrew}{ אני אוהב חיות } &
 7\%         & content    \\
\texttt{wrong language}     & \begin{tabular}[c]{@{}c@{}}``source'': ``\foreignlanguage{arabic}{ هذا الصباح انت حزين , تريد أن تستمع   لاغنيتك }'', ``target'': ``\foreignlanguage{chinese}{今天早上你很伤心，你想听你的歌}'', \\ ``translation'': \textcolor{red}{``This morning you are sad, you want to listen to your song''}\end{tabular}                                                                    & 7\%         & content    \\
\texttt{language prefix}    &      \textcolor{red}{Arabic:}\foreignlanguage{arabic}{ النص المترجم هنا
} & 7\%         & formatting \\
\texttt{extra punctuation}  & \foreignlanguage{chinese}{我爱你!}\textcolor{red}{!!!}                                                                                                                                        & 3\%         & formatting \\
\texttt{code block}         & \textcolor{red}{\textasciigrave \textasciigrave \textasciigrave }\foreignlanguage{chinese}{你好}\textcolor{red}{\textasciigrave \textasciigrave \textasciigrave }       & 3\%         & formatting \\
\texttt{special formatting} & \textcolor{red}{{[}{[}}\textbackslash{}``Australian Shepherd\textbackslash{}''\textcolor{red}{{]}{]}}                                                                                                                                                                                                               & 2\%         & formatting \\
\texttt{translation prefix} & \textcolor{red}{Translation:} \foreignlanguage{chinese}{八项维和任务}                                                                                                                                                                                                                                                                & 1\%         & formatting \\
\texttt{cultural note}      & \begin{tabular}[c]{@{}c@{}}\foreignlanguage{chinese}{请听葬礼协奏曲}
\textcolor{red}{\textbackslash{}n\textbackslash{}n(Note: The `Funeral Concerto' is not a specific well-known piece by a single composer} \\ \textcolor{red}{— this may refer to Tchaikovsky's Piano   Concerto No. 1...}\end{tabular}                                               & 1\%         & content   \\ \bottomrule
\end{tabular}%
}
\caption{Noise patterns with examples and frequencies (\%) in the dataset detected by Claude Opus 4.6. We use ``...'' to denote the omitted long outputs. Text in red denotes noise.}
\label{tab:noise_pattern}
\end{table*}

\subsection{Noise Rate} \label{sec:noise_rate}

To estimate how frequently translation noise appears in LLM outputs, we employ a lightweight rule-based detector that captures two common types of noise: explanatory text and text generated in the wrong target language. The detector is intended to provide a coarse estimate of noise prevalence rather than a complete characterization of all noise types. The noise rate (Noise\%) for a given prompt is formally defined as:

\begin{equation} \label{eq:noise}
    \mathrm{Noise}\% = \frac{|E \cup W|}{N}
\end{equation}

\noindent where $N$ denotes the total number of translation instances, $E$ represents the set of outputs containing explanatory text, and $W$ denotes the set of outputs containing text in the wrong target language. We further define $\mathrm{Exp}\% = \frac{|E|}{N}$ and $\mathrm{WrongL}\% = \frac{|W|}{N}$ to separately measure the proportions of explanatory noise and wrong-language outputs.

Explanatory text is detected using regular expressions matching common explanatory markers (e.g., ``explanation'' and similar meta-linguistic phrases in Appendix~\ref{appendix_regex}). Wrong-language outputs are identified using the fastText language identification model \citep{bojanowski-etal-2017-enriching}, with a confidence threshold of 60\%. An output is considered noisy if either type of signal is detected. While this rule-based detector does not capture all possible noise types, it is sufficient, as evaluated in $\S$\ref{sec:results}, for estimating noise frequencies across prompts and models, and identifying noisy candidates for human curation in $\S$\ref{sec:authetic_curation}.

Table~\ref{tab:clean_rate} reports Noise\% across prompts and models. Prompt 0 produces the highest proportion of noisy outputs among the three prompts, particularly for wrong target-language generation, and we therefore use its outputs for the subsequent noise analysis in $\S$\ref{sec:noise_analysis}. Among models, \Gemma exhibits the highest noise rates (up to 99.73\%); manual inspection confirms that it frequently generates explanatory text alongside translations. These high noise rates across models and prompts underscore the need for a systematic study of this problem and for dedicated methods to extract clean translations for fair MT evaluation.

\subsection{Noise Analysis} \label{sec:noise_analysis}

We utilize all 792,000 translation outputs generated by 12 models with Prompt 0 across 22 language pairs for noise analysis. To identify recurring noise patterns in such a large dataset, we conduct a two-stage analysis. First, we use an LLM, Claude Opus 4.6 \citep{claude2026}, to assist in summarizing and grouping similar noise behaviors across the outputs. Given a generated translation, the model is prompted to propose representative noise patterns, estimate their relative frequencies, and extract up to ten representative examples for each pattern\footnote{All instances are saved when fewer than ten are available.}. The identified patterns and examples (see Table~\ref{tab:noise_pattern}) are subsequently verified and refined through manual inspection. 

Based on this inspection, we categorize the twelve observed noise patterns into two groups: \textit{content} and \textit{formatting} noise, corresponding to semantic and presentation-level artifacts, respectively. We retain a relatively fine-grained set of patterns to facilitate synthetic noise generation, while noting that alternative taxonomies are also possible. Since manually validating the exact frequency of each pattern at this scale is infeasible, the reported frequencies should be interpreted as approximate estimates rather than precise measurements.\footnote{To assess their reliability, we independently reproduce the frequency estimation using gemma-4-31B-it \citep{gemma4blog}, which yields a strong and statistically significant positive rank correlation with Claude Opus 4.6 (Spearman's $\rho=0.84$).} These estimated frequencies, together with the taxonomy, are used primarily to capture the overall distribution of noise patterns and to support the generation of synthetic noise that reflects realistic noise distributions, while the representative examples are used as few-shot demonstrations for synthetic noise generation in $\S$\ref{sec:synthetic_gen}.

As shown in Table~\ref{tab:noise_pattern}, \texttt{explanation} accounts for the largest share of noisy outputs (\textasciitilde 33\%), primarily produced by \Gemma and DeepSeek-V3.2-Exp (see Table~\ref{tab:clean_rate}). Other frequent patterns include \texttt{alternative} translations and \texttt{off-topic} responses. Generation in the wrong target language\footnote{Manual inspection of saved samples suggests they are mostly in English or in the source language.} also contributes a notable proportion (\textasciitilde 7\%). Overall, content-level noise constitutes the majority of noisy translations and is generally more challenging to handle than \textit{formatting} noise.

\section{Benchmark Construction} \label{sec:bench_build}

To facilitate systematic research on clean translation detection and extraction, we introduce \textbf{TransClean}, the first benchmark designed to evaluate methods for removing noise from LLM-generated translations. It comprises 9,900 paired instances of LLM generated translations and their clean translation counterparts. Although the translations generated in \S\ref{sec:trans_noise} contain various levels of noise, their direct use would require manual annotation of the clean translation, which would be prohibitively expensive and require annotators with expertise in many languages. Therefore, we adopt a hybrid strategy instead: we generate large-scale \textit{synthetic} noisy translations using LLMs, guided by the empirically observed noise patterns described in $\S$\ref{sec:noise_analysis}. To further validate the realism and usefulness of the synthetic data, we additionally curate a smaller subset of \textit{authentic} noisy translations paired with silver clean translations. This subset enables comparison between synthetic and real noise scenarios. The construction of the synthetic dataset and the curated subset are described in $\S$\ref{sec:synthetic_gen} and $\S$\ref{sec:curate_noise} respectively, while representative examples of the final constructed subsets are provided in Appendix~\ref{appendix_examples}.

\subsection{Synthetic Noise Generation} \label{sec:synthetic_gen}

Because the noise observed in LLM translation outputs originates from LLM generation behaviors, we use LLMs to simulate these noise patterns. Synthetic noisy translations are generated by injecting noise patterns (Table~\ref{tab:noise_pattern}) into reference translations, which serve as clean ground-truth translations.

\paragraph{Noise Generation} We first sample 100 instances for each LP whose source texts contain at least 10 words\footnote{Words are defined as space-separated units for languages using whitespace. For languages without whitespace segmentation, characters are counted instead.}. Their reference translations are treated as clean translations, yielding 2,200 clean instances across the 22 LPs. Based on these clean translations, we generate noisy outputs under 3 noise categories: \textit{content}, \textit{formatting}, and their combination (\textit{combo}). Each category contains 2,200 instances. The specific noise pattern applied to each instance is sampled according to the empirical distribution observed in Table~\ref{tab:noise_pattern}. As a result, the distribution of noise patterns in the synthetic data approximately matches the distribution observed in LLM outputs.

In practice, noise patterns are sampled using their empirical frequencies as weights. For \textit{formatting} noise, including \texttt{language prefix}, \texttt{translation prefix}, \texttt{extra punctuation}, \texttt{code block}, and \texttt{special formatting}, we apply a rule-based generator that inserts formatting artifacts into the reference translation. For the remaining patterns, including all \textit{content} patterns and the \textit{formatting} pattern \texttt{verbose preamble}, we use GPT-5-mini \citep{Singh2026-zj} to generate noisy translations in a few-shot prompting setup. The demonstrations consist of representative examples extracted during the noise analysis stage ($\S$\ref{sec:noise_analysis}). The prompt template is provided in Appendix~\ref{appendix_noise_creation}. For all noise patterns, the reference translation is used as the gold clean translation, except \texttt{off-topic} and \texttt{wrong language}, whose gold clean is an empty string.

Overall, the synthetic dataset contains 8,800 instances distributed across four splits: three noisy categories and one clean category. The clean split serves as a control set to evaluate whether extraction methods preserve already clean translations. 
Detailed statistics of the synthetic dataset are shown in Table~\ref{tab:synthetic_stats}.

\paragraph{Manual Validation} To assess the realism and correctness of the generated noise, we conduct manual validation on a subset of the synthetic data. Specifically, we randomly sample 10 instances for each of the seven LLM-generated noise patterns (i.e., \texttt{explanation}, \texttt{alternatives}, \texttt{off-topic}, \texttt{verbose preamble}, \texttt{bilingual output}, \texttt{wrong language}, and \texttt{cultural note}). For the \textit{combo} category, we additionally sample 30 instances. This results in 100 manually inspected instances covering 9 language pairs. Manual inspection confirms that the generated outputs correctly reflect the intended noise patterns and closely resemble the noise behaviors in real LLM translations.\footnote{In the case of \texttt{wrong language}, we found that the LLM typically generates text in English or in the source language, which reflects the type of language confusion found in the noise analysis in $\S$~\ref{sec:noise_analysis}.}

\subsection{Curated Noisy Subset} \label{sec:curate_noise}

To complement the synthetic dataset, we construct a curated subset of authentic noisy translations drawn from real LLM outputs. The curated subset contains 1,100 instances, each annotated with a noise pattern and a silver clean translation.

\subsubsection{Authentic Noise Curation} \label{sec:authetic_curation}

Taking the 792,000 LLM translation outputs from Prompt 0 in \S\ref{sec:data_src} as a starting point, we first apply the rule-based detector described in $\S$\ref{sec:noise_rate} to filter potentially noisy outputs, yielding 232,403 candidate instances. We then use GPT-5-mini to identify authentic noisy translations among the candidate instances. For each instance classified as noisy, the model assigns a noise pattern label. We curate 50 instances per language pair, each annotated with its corresponding noise pattern. We then verify the detected instances to confirm they represent authentic noise. The prompt used for this task is provided in Appendix~\ref{appendix_curation}.

\subsubsection{Clean Translation Annotation} \label{sec:clean_anno}

We employ three LLMs, GPT-5-mini, Qwen3.5-122B-A10B \citep{qwen35blog}, and gemma-4-31B-it to annotate the clean translation for each noisy output. The models are provided with the noisy translation and its corresponding noise label as context. The prompt used for this task is shown in Appendix~\ref{appendix_anno}.

\begin{table}[h]
\centering
\begin{tabular}{ccc}
\toprule
Agreement      & Instances & Percentage \\ \midrule
3/3  & 448       & 40.73\%    \\
2/3  & 337       & 30.64\%    \\
Tie & 315       & 28.64\%    \\ \bottomrule
\end{tabular}%
\caption{Agreement of the three LLMs on annotating clean translations for the 1100 curated examples.}
\label{tab:agreements}
\end{table}

We adopt a majority voting strategy to determine the final silver clean translation. If at least two models produce identical outputs, the shared translation is used as the label. In 48 cases where at least one model outputs an empty string, manual inspection confirms that the corresponding outputs are \textit{off-topic} responses, and the empty string is therefore retained as the correct label. Agreement statistics of the three models are presented in Table~\ref{tab:agreements}. Among the remaining 315 instances without majority agreement, we manually examine 143 instances for which the authors are fluent speakers of the target language. In these cases, gemma-4-31B-it produces the correct clean translation for all instances except 17 where multiple valid translations exist and all model outputs are acceptable. Based on this observation, we adopt the output of gemma-4-31B-it as the silver label for the remaining disagreement cases.

\section{Translation Extraction} \label{sec:trans_ext}

To support fair MT evaluation beyond merely detecting noise, we propose two methods that can extract clean translations from noisy LLM outputs: a span-based method using quality estimation models in $\S$\ref{baseline}, and an LLM-based extraction method in $\S$\ref{llm_extractor}. Evaluation metrics and results are presented in $\S$\ref{sec:eval_metrics} and $\S$\ref{sec:results}. The rule-based detector is evaluated in $\S$\ref{sec:results} for noise detection only to compare with the proposed methods. Details for running these extraction methods are in Appendix \ref{appendix_ext_details}. 

\subsection{Span-based Extraction} \label{baseline}

The observation that lengthy explanatory text constitutes the largest source of noise in LLM outputs, and that explanations are typically separated from the translation by line breaks, motivates a span-based approach: splitting the output into shorter spans, identifying the span most likely to contain the clean translation, and removing any residual noise from that span.

Following common patterns observed in LLM-generated text, we segment the output using the line feed character (\texttt{\textbackslash n}) as a delimiter, yielding a set of candidate spans $S=\{s_1, s_2, \dots, s_m\}$. We then score each span using COMET-KIWI \citep{rei-etal-2022-cometkiwi}, a reference-free quality estimation (QE) model, which produces a score $q(s_j, x_{\text{src}})$ by comparing each span $s_j$ against the source text $x_{\text{src}}$. The candidate clean translation is selected as:

\begin{equation}
s^* = 
\begin{cases} 
s_1, & \text{if } |S| = 1 \\ 
\displaystyle\arg\max_{s_j \in S} \, q(s_j, x_{\text{src}}), & \text{if } |S| > 1 
\end{cases}
\end{equation}

\noindent That is, if the output contains a single span, it is directly taken as the candidate translation without QE scoring. Otherwise, the span with the highest QE score is selected.

Finally, a rule-based post-processing step is applied to $s^*$ to remove any residual formatting noise, such as language prefixes, yielding the extracted translation $\hat{t} = g(s^*)$, where $g(\cdot)$ denotes the rule-based cleaning function.

\subsection{LLM-based Extraction} \label{llm_extractor}

As an alternative to the span-based method, we propose using LLMs directly as extractors to produce clean translations. Given a noisy LLM translation output $x_i$, the extraction is formulated as:

\begin{equation}
\hat{t}_i = \mathcal{M}_{\text{ext}}([p; x_i])
\end{equation}

\noindent where $\mathcal{M}_{\text{ext}}$ is the extractor LLM and $p$ is a fixed prompt template instructing the model to extract the clean translation from $x_i$ (see Appendix~\ref{appendix_extractor} for the full template). Notably, the extractor receives only the LLM translation output $x_i$. No reference translation, or description of noise patterns is provided. This constraint ensures a fair comparison with the span-based approach, which likewise operates without access to reference information.

This setup also distinguishes this LLM extraction approach from the silver clean translation annotation procedure described in $\S$\ref{sec:clean_anno}, where the annotator LLM is given both the source text and reference translations, along with explicit noise pattern descriptions.

In practice, we employ two backbone extractors under a zero-shot setting: a multilingual dense LLM, \Aya, and Qwen3.5-122B-A10B, an English- and Chinese-dominant mixture-of-experts model.

\subsection{Evaluation Metrics} \label{sec:eval_metrics}

To evaluate how effectively our methods detect noise and extract clean translations from LLM outputs with our benchmark, we introduce two metrics: \textit{detection accuracy} and \textit{extraction accuracy}.

\paragraph{Detection Accuracy}
Detection accuracy measures the rate at which a method correctly identifies whether an LLM translation output is noisy or clean (i.e., translation-only). Given the $i$-th LLM translation output $x_i$ and an extraction method $f(\cdot)$, the predicted label is determined by:
\begin{equation}
    \hat{y}_i = 
    \begin{cases} 
        0 \ (\text{clean}), & \text{if } f(x_i) = x_i \\ 
        1 \ (\text{noisy}), & \text{if } f(x_i) \neq x_i 
    \end{cases}
\end{equation}
That is, if the extraction method returns the input unchanged, the sample is classified as clean; any modification to the input implies the presence of noise. Detection accuracy ($\text{Acc}\textsubscript{det}$) is computed as the proportion of samples for which the predicted noise label $\hat{y}_i $ matches the ground-truth label $y_i$.

%
%

\paragraph{Extraction Accuracy}
Extraction accuracy is a stricter metric that measures whether the extracted translation exactly matches the annotated clean reference. Normalization is applied to both strings prior to comparison. It is formally defined as:
\begin{equation}
    \text{Acc}_{\text{ext}} = \frac{1}{N} \sum_{i=1}^{N} \mathbbm{1}\!\left[\texttt{norm}(\hat{t}_i) = \texttt{norm}(t_i)\right]
\end{equation}
where $\hat{t}_i$ is the extracted translation for the $i$-th sample, $t_i$ is the corresponding annotated clean reference, $\mathbbm{1}[\cdot]$ is the indicator function, $N$ is the total number of samples, and $\texttt{norm}(\cdot)$ denotes the normalization function applied before comparison. Specifically, $\texttt{norm}(\cdot)$ strips leading and trailing whitespace and applies Unicode NFC normalization to $\hat{t}_i$ and $t_i$.

\subsection{Evaluation Results} \label{sec:results}

\begin{table}[ht]
\centering
\resizebox{7.5cm}{!}{%
\begin{tabular}{lcccc}
\toprule
                & \multicolumn{2}{c}{Synthetic} & \multicolumn{2}{c}{Curated} \\
Method         & $\text{Acc}_{\text{det}}$      & $\text{Acc}_{\text{ext}}$      & $\text{Acc}_{\text{det}}$     & $\text{Acc}_{\text{ext}}$     \\ \midrule
Span-based     & 99.68         & 50.59         & 96.36        & 15.82        \\
Qwen extractor & 98.38         & \textbf{54.07}         & 96.73        & \textbf{52.18}        \\
Aya extractor  & 89.94         & 36.69         & 99.73        & 51.36     \\ 
Rule-based detector & 87.26 & / & 98.27 & / \\
\bottomrule
\end{tabular}%
}
\caption{Detection and extraction accuracy (\%) on the synthetic and curated subsets of our benchmark.}
\label{tab:results_ext}
\end{table}

\begin{table*}[ht]
\centering
\resizebox{11cm}{!}{%
\begin{tabular}{ccccccc}
\toprule
Method         & Formatting & Content & Combo & Clean & Noisy & Overall \\ \midrule
Span-based     & 83.68      & 10.41   & 8.27 & \textbf{100}   & 34.12 & 50.59 \\
Qwen extractor & \textbf{89.91}   & \textbf{17.68}   & \textbf{14.95} & 93.73 & \textbf{40.85} & \textbf{54.07} \\
Aya extractor  & 57.91      & 14.50   & 14.45 & 59.91 & 28.95 & 36.69 \\
\bottomrule
\end{tabular}%
}
\caption{Extraction accuracy (\%) for each noise split (category) of the synthetic subset. The ``noisy'' column is the combination of the first three categories.}
\label{tab:acc_category}
\end{table*}

\paragraph{Overall Results} Table \ref{tab:results_ext} reports the detection and extraction accuracy on both the synthetic and curated subsets. Detection accuracy is nearly saturated (close to 100\%) for most methods and both subsets, indicating that identifying whether a translation contains noise is relatively easy. Our rule-based detector performs well, especially on the curated noisy subset, demonstrating its effectiveness in detecting noise and calculating Noise\%.

Extraction, however, remains challenging. The best-performing method, \textit{Qwen extractor}, achieves 54.07\% accuracy on the synthetic subset and 52.18\% on the curated subset. While promising under strict exact-match evaluation, these results suggest substantial room for improvement in extracting clean translations from noisy outputs.

The span-based approach performs well on the synthetic dataset but drops sharply on the curated subset for extraction accuracy. This behavior is expected: the synthetic data follows predefined noise patterns aligned with the rule-based cleaning function, whereas the curated subset contains authentic noise that may not match these patterns. In contrast, LLM-based extraction approaches exhibit more stability across datasets, suggesting better generalization to diverse noise patterns.

An exception is \textit{Aya extractor}, whose accuracy increases on the curated subset while most other methods decline. To understand this behavior, we analyze its performance on each split of the synthetic subset. Aya achieves only 59.91\% accuracy on the clean split, substantially lower than other methods (above 90\%). Inspection shows that Aya frequently paraphrases already clean translations—about 40\% of the time (892/2200)—altering wording, punctuation, or sentence structure. These minor reformulations lead to mismatches under exact-match evaluation and largely explain its lower synthetic-set performance.

Overall, the relative performance trends are consistent across synthetic and curated subsets, suggesting that the synthetic data reasonably approximates real noisy translations and serves as a reliable benchmark to evaluate extraction methods.

\paragraph{Results per Noise Split} Table \ref{tab:acc_category} presents the extraction accuracy for each noise split of the synthetic subset. \textit{Qwen extractor} achieves the best performance across all noise categories. The only exception is the clean split, where the span-based method attains 100\% accuracy by preserving the original translation when no noise is detected.

Across noise categories, \textit{formatting} noise yields the highest accuracy among the noisy splits. This is expected, as formatting noise can often be removed with simple transformations. The most difficult category is \textit{combo}, which combines \textit{content} and \textit{formatting} noise. The interaction of multiple noise types substantially increases extraction difficulty, leading to lower accuracy for all methods.

These results further support the design of the synthetic benchmark: the splits exhibit distinct difficulty levels and capture meaningful differences among noise categories, enabling more fine-grained evaluation of extraction approaches.

\section{Related Work} \label{sec:background}

LLMs are increasingly used to generate synthetic data for Natural Language Processing tasks due to their strong language modeling and controllable generation capabilities \citep{long-etal-2024-llms,Andreea2025}. In MT, synthetic data has long been used through techniques such as back-translation to improve performance, particularly for low-resource languages \citep{hassan-etal-2017-synthetic,poncelas-etal-2018-investigating}. More recent work employs LLMs directly to generate synthetic multilingual data. For example, \citet{de-gibert-etal-2025-scaling} generate translations for several low-resource languages using GPT-4o \citep{OpenAI2024-hl} and show that synthetic data can improve downstream MT systems despite its noise. However, prior work primarily uses synthetic data to improve translation models rather than to study the behavior of LLM-generated translations themselves. In this work, we instead leverage LLMs to generate synthetic translation noise based on empirically observed patterns, enabling scalable construction of a benchmark for clean translation extraction.


\section{Conclusion} \label{sec:conclusion}

In this work, we present a systematic study of \textit{translation noise}. Through large-scale analysis of more than 790,000 LLM translation outputs across 22 language pairs, we identify 12 recurring noise patterns and categorize them into \textit{formatting} and \textit{content} noise. Based on these observations, we introduce \textbf{TransClean}, the first benchmark designed to evaluate methods that extract clean translations from noisy LLM outputs. The benchmark combines a large synthetic dataset with gold clean translations and a curated subset of authentic noisy translations with silver clean translation, enabling both controlled evaluation and validation on realistic data. Using this benchmark, we evaluate span-based and LLM-based extraction approaches and show that, while noise detection is relatively straightforward, clean translation extraction remains a challenging task with substantial room for improvement.

In future work, we plan to develop more robust methods for clean translation extraction and explore approaches that better generalize to diverse noise patterns across languages and models. We hope that TransClean will facilitate further research toward more reliable use of LLMs for translation and other structured generation tasks.

\section*{Limitations}

This work has several limitations. First, our estimation of the noise rate relies on a coarse rule-based detector that identifies explanatory text through English keyword matching and detects wrong-language outputs using automatic language identification. While this approach enables scalable analysis across hundreds of thousands of translation outputs, it may miss some noise instances that do not match the predefined patterns or may occasionally produce false positives. Developing more reliable detection methods is therefore an important direction for future work. One motivation of TransClean is precisely to provide a benchmark that enables systematic evaluation of improved detection and extraction approaches.

Second, the identification of noise patterns was assisted by an LLM due to the scale of the collected outputs (over 790,000 translations), which makes full manual inspection impractical. Although we subsequently verified the discovered patterns and examples through manual review, the taxonomy of noise patterns may not be exhaustive and could evolve as new models or prompting strategies produce different types of noise.

Third, the synthetic noise of our benchmark is generated based on observed patterns and their empirical distribution. While this design enables controlled evaluation and sufficient scale, synthetic noise may not fully capture the diversity and complexity of noise produced by LLMs in real-world settings. In addition, although we include a curated subset of authentic noise, the benchmark remains largely English-centric because both the translation prompts and the noise-generation prompts are written in English. As a result, the generated noise may under-represent truly multilingual or language-specific noise phenomena. Extending the benchmark with more diverse multilingual noise patterns remains an important direction for future work.

Finally, exact-match extraction accuracy may be overly stringent as the sole primary metric. We observe that it can penalize semantically correct outputs when Aya paraphrases translations that are already clean. This highlights a potential mismatch between exact-match evaluation and the semantic correctness of the extracted translations. Softer edit-based measures or semantic similarity metrics could therefore provide a more informative complement to exact-match accuracy in future work. 

\section*{Ethical Considerations}

This research relies exclusively on publicly accessible datasets, with all data utilization adhering to the licensing agreements specified by \citet{Kulkarni2015}, \citet{scherrer-etal-2014-swissadmin}, \citet{Park2019-vd}, and \citet{barrault-etal-2020-findings}. It is presumed that these repositories contain no sensitive or personally identifiable information. Consequently, their application in this study is deemed to present no significant ethical risks. Furthermore, the systematic generation of synthetic noise and the curation of authentic noise samples are not expected to yield additional personal data or introduce further ethical complications. In the interest of transparency and reproducibility, the resulting dataset is released to the public domain. 

All original ideas, analyses, and content in this paper were created by the authors. AI tools were used only as supportive aids for improving writing quality and assisting with coding tasks. The authors retain full responsibility for the intellectual content, analyses, and conclusions presented in this work.

\section*{Acknowledgments}

This work has received funding from the European Union's Horizon Europe research and innovation programme under the Marie Skłodowska-Curie grant agreement No. 101126636. 

The computations were performed on resources provided through Sigma2---the national research infrastructure provider for high-performance computing and large-scale data storage in Norway. We acknowledge Norway and Sigma2 for awarding this project access to the Olivia supercomputer, through Project nn9851k.

\bibliography{custom,anthology-1,anthology-2}



\appendix
\counterwithin{figure}{section}
\setcounter{figure}{0} 
\counterwithin{table}{section}
\setcounter{table}{0} 

\section{Appendix: Additional Tables for Data and Models} \label{appendix_add_tables}

\begin{table*}[h]
\centering
\resizebox{10cm}{!}{%
\begin{tabular}{ccc}
\toprule
Lang\_pairs             & Test\_size & Source                            \\ \midrule
Arabic-Chinese (ar-zh)  & 3,000      & TED Multilingual Parallel Corpora \\
Arabic-Hebrew (ar-he)   & 3,000      & TED Multilingual Parallel Corpora \\
Chinese-French (zh-fr)  & 3,000      & TED Multilingual Parallel Corpora \\
Chinese-Russian (zh-ru) & 3,000      & TED Multilingual Parallel Corpora \\
French-Italian (fr-it)  & 3,000      & SwissAdmin                        \\
German-French (de-fr)   & 3,000      & SwissAdmin                        \\
German-Italian (de-it)  & 3,000      & SwissAdmin                        \\
Korean-Chinese (ko-zh)  & 3,000      & Chinese-Korean Parallel Corpus    \\
Korean-French (ko-fr)   & 3,000      & TED Multilingual Parallel Corpora \\
Russian-French (ru-fr)  & 3,000      & TED Multilingual Parallel Corpora \\
English-Chinese (en-zh) & 3,000      & WMT20 QE Shared Task              \\
English-Czech (en-cs)   & 3,000      & WMT20 QE Shared Task              \\
English-German (en-de)  & 3,000      & WMT20 QE Shared Task              \\
English-Polish (en-pl)  & 3,000      & WMT20 QE Shared Task              \\
English-Russian (en-ru) & 3,000      & WMT20 QE Shared Task              \\
English-Tamil (en-ta)   & 3,000      & WMT20 QE Shared Task              \\
Chinese-English (zh-en) & 3,000      & WMT20 QE Shared Task              \\
Czech-English (cs-en)   & 3,000      & WMT20 QE Shared Task              \\
German-English (de-en)  & 3,000      & WMT20 QE Shared Task              \\
Khmer-English (km-en)   & 3,000      & WMT20 QE Shared Task              \\
Russian-English (ru-en) & 3,000      & WMT20 QE Shared Task              \\
Tamil-English (ta-en)   & 3,000      & WMT20 QE Shared Task              \\ \bottomrule
\end{tabular}%
}
\caption{The size of our test set for each language pair and their corresponding sources.}
\label{tab:data_details}
\end{table*}

\begin{table*}[ht]
\centering
\resizebox{15cm}{!}{%
\begin{tabular}{cccc}
\toprule
Model Name                              & Architecture          & Instruction-tuned or Reasoning & Parameter Size         \\ \midrule
\QwenLargeInstruct         & decoder-only-moe      & instruction-tuned                       & 30B in total, 3B active \\
\QwenLargeThink         & decoder-only-moe      & reasoning                                & 30B in total, 3B active \\
\QwenSmallInstruct              & decoder-only-dense    & instruction-tuned                        & 4B                      \\
\QwenSmallThink              & decoder-only-dense    & reasoning                               & 4B                      \\
\Llama         & decoder-only-dense    & instruction-tuned                        & 3B                      \\
\Gemma                    & decoder-only-dense    & instruction-tuned                       & 27B                     \\
\QwenOld                & decoder-only-dense    & instruction-tuned                       & 32B                     \\
\DeepseekQwen  & decoder-only-dense  & reasoning                                & 32B                     \\
\Aya               & decoder-only-dense    & instruction-tuned                        & 32B                     \\
\Tower                   & decoder-only-dense    & instruction-tuned                        & 72B                     \\
\TFiveGemma         & encoder-decoder-dense & instruction-tuned                        & 4B                      \\
DeepSeek-V3.2-Exp      & decoder-only-moe      & mixed                                   & 671B in total, 37B active         \\      
\bottomrule
\end{tabular}%
}
\caption{Model details including names, architectures, size and either instruction-tuned or reasoning variants.}
\label{tab:model_details}
\end{table*}

\section{Appendix: LLM Inference Details} \label{sec:appendix_inf}

We used vLLM \cite{Kwonetal2023} for inference with most models with the exception of DeepSeek-V3.2-Exp \citep{DeepSeek-AI2025-hr} and \TFiveGemma. For these models, we obtained inference results using the respective API or the HuggingFace Transformers library \cite{wolf-etal-2020-transformers}. We kept the default values of the hyperparameters with temperature and top\_p both set to 1. With the exception of DeepSeek-V3.2-Exp, all models were run without quantization on 4 NVIDIA GH200 GPUs. On average, an instruction-tuned model requires approximately 10 minutes to process one language pair (3,000 instances), whereas a reasoning model requires about 18 minutes. For reasoning LLMs, only content after the reasoning tags (i.e.,<think></think>) is treated as ``LLM outputs'' for noise analysis.

\section{Appendix: Regular Expressions for Rule-based Detector} \label{appendix_regex}

We use the following regular expression patterns to detect explanatory or meta-linguistic content.

\subsection{Common Explanation Phrases}

\begin{lstlisting}[language=Python]
    r'\b(the translation is|here is|here\'s|this translates to|translation:|translated text:|output:|target:)\b',
    r'\b(in \w+ (this|it) (means|says|translates))\b',
    r'\b(note that|please note|it should be noted)\b',
    r'\b(explanation|reasoning|analysis|breakdown)\b',
    r'\n\s*(translation|explanation|note|original|source|target)\s*:',
\end{lstlisting}

\subsection{Meta-linguistic Markers}

\begin{lstlisting}[language=Python]
    r'\b(literally|figuratively|idiomatically|contextually)\b',
    r'\b(this (word|phrase|sentence|text))\b',
    r'\b(means|refers to|indicates|suggests)\b.*\b(that|which)\b',
\end{lstlisting}

\subsection{Comments to user}

\begin{lstlisting}[language=Python]
    r'\b(hope this helps|let me know|feel free|if you|you can)\b',
    r'\b(please|kindly|note:|important:)\b',
\end{lstlisting}

\subsection{Thinking Markers}

\begin{lstlisting}[language=Python]
    r'<think>|</think>|<thought>|</thought>',
    r'\*\*reasoning\*\*|\*\*analysis\*\*|\*\*explanation\*\*',
\end{lstlisting}

\subsection{Markdown or XML Tags}
\begin{lstlisting}[language=Python]
    r'^#+\s+',
    r'<[a-zA-Z]+>.*</[a-zA-Z]+>',
\end{lstlisting}

\subsection{Lists or Parenthetical Explanations}

\begin{lstlisting}[language=Python]
    r'^\s*[\d\-\*]+[\.\)]\s+',
    r'\([^)]{50,}\)'
\end{lstlisting}

\section{Appendix: Examples of TransClean} \label{appendix_examples}

\subsection{An Example of the Synthetic Subset}

\begin{Verbatim}[formatcom=\small,
    breaklines=true,       % Enables automatic line breaking
    breakanywhere=true,
    breakbytoken=false ]
    "source": "Die Wasserqualität hat sich in den letzten Jahrzehnten deutlich verbessert.", 
    "translation": "The water quality has greatly improved over the past decades.", 
    "src_lang": "de", "tgt_lang": "fr", 
    "noise_pattern": "wrong_language",
    "gold_reference": "La qualité de l’eau s’est sensiblement améliorée au cours des dernières décennies."
\end{Verbatim}

\subsection{An Example of the Curated Subset}

\begin{Verbatim}[formatcom=\small,
    breaklines=true,       % Enables automatic line breaking
    breakanywhere=true,
    breakbytoken=false,
    commandchars=\@\{\}
]
    "source": "Dog control laws to be reviewed in government consultation",
    "translation": "English: Dog control laws to be reviewed in government consultation  \nChinese: @cjk{政府咨询将审查狗只控制法例}",
    "src_lang": "en", "tgt_lang": "zh",
    "noise_patterns": ["language_prefix", "bilingual_output"], "primary_pattern": "bilingual_output",
    "silver_reference": "@cjk{政府咨询将审查狗只控制法例}", "silver_agreement": 3, "silver_votes": ["@cjk{政府咨询将审查狗只控制法例}", "@cjk{政府咨询将审查狗只控制法例}", "@cjk{政府咨询将审查狗只控制法例}"]
\end{Verbatim}

\section{Appendix: Prompt for Synthetic Noise Generation} \label{appendix_noise_creation}

\begin{figure*}[ht]
\centering
\fbox{%
  \begin{minipage}{16cm}
    \textbf{SYSTEM PROMPT} \\
    You are simulating a large language model that generates translations with extra noise, explanations, or formatting artifacts. Your task is to take a clean reference translation and add realistic noise to it — exactly as a helpful-but-verbose LLM would. \\
    Rules: \\
    - Output ONLY the noisy translation (no meta-commentary, no JSON). \\
    - The core translation meaning must remain correct. \\
    - The noise must look authentic, as if a real LLM produced it. \\
    - Use the target language for the translation itself; explanatory text may be in English or the source/target language depending on the pattern.

    \textbf{USER PROMPT} \\
    \{NOISE PATTERN DESCRIPTION\} \\
    \{FEW-SHOT EXAMPLES\} \\
    Source text: \{source\} \\
    Source language: \{src\_lang\_name\} \\
    Target language: \{tgt\_lang\_name\} \\
    Clean translation: \{clean\_translation\} \\
    Generate the noisy output \{BASED ON NOISE DESCRIPTION\}:
  \end{minipage}
  }
  \caption{Prompt for generating synthetic noise.}
\label{synthetic_noise}
\end{figure*}

\section{Appendix: Statistics of the Synthetic Subset} \label{appendix_synt_stats}

\begin{table*}[h]
\centering
\begin{minipage}[t]{0.45\textwidth}
    \centering
    \resizebox{6.5cm}{!}{%
    \begin{tabular}{ccc}
    \toprule
    Noise pattern       & Count & Generation method \\ \midrule
    \texttt{explanation}         & 965   & GPT-5-mini        \\
    \texttt{verbose preamble}   & 791   & GPT-5-mini        \\
    \texttt{language prefix}    & 650   & rule-based        \\
    \texttt{alternatives}        & 460   & GPT-5-mini        \\
    \texttt{off-topic}          & 320   & GPT-5-mini        \\
    \texttt{code block}         & 267   & rule-based        \\
    \texttt{extra punctuation}  & 244   & rule-based        \\
    \texttt{bilingual output}   & 216   & GPT-5-mini        \\
    \texttt{wrong language}     & 209   & GPT-5-mini        \\
    \texttt{special formatting} & 164   & rule-based        \\
    \texttt{translation prefix} & 84    & rule-based        \\
    \texttt{cultural note}      & 30    & GPT-5-mini        \\
    Subtotal            & 4,400 & /                 \\
    \bottomrule
    \end{tabular}%
    }
    \subcaption{Counts per noise pattern}
    \label{tab:N_patterns}
\end{minipage}%
\hspace{1cm}
\begin{minipage}[t]{0.45\textwidth}
    \centering
    \resizebox{6.5cm}{!}{%
    \begin{tabular}{cc}
    \toprule
    Combo patterns                        & Count \\ \midrule
    \texttt{explanation + language prefix}           & 644   \\
    \texttt{alternatives + language prefix}          & 344   \\
    \texttt{off-topic + language prefix}            & 249   \\
    \texttt{explanation + verbose preamble}          & 220   \\
    \texttt{bilingual output + language prefix}     & 161   \\
    \texttt{wrong language + language prefix}       & 137   \\
    \texttt{alternatives + verbose preamble}        & 96    \\
    \texttt{off-topic + verbose preamble}           & 76    \\
    \texttt{explanation + translation prefix}        & 58    \\
    \texttt{wrong language + verbose preamble}      & 47    \\
    \texttt{bilingual output + verbose preamble}    & 45    \\
    \texttt{alternatives + translation prefix}      & 40    \\
    \texttt{cultural note + language prefix}        & 33    \\
    \texttt{off-topic + translation prefix}         & 18    \\
    \texttt{wrong language + translation prefix}    & 12    \\
    \texttt{bilingual output + translation prefix}  & 9     \\
    \texttt{cultural note + verbose preamble}       & 9     \\
    \texttt{cultural note + translation prefix}     & 2     \\ 
    Subtotal                              & 2,200 \\
    \bottomrule
    \end{tabular}%
    }
    \subcaption{Combo pattern counts}
    \label{tab:combo_patterns}
\end{minipage}
\caption{Detailed statistics of the synthetic noise data: (a) counts per noise pattern and (b) combo pattern counts.}
\label{tab:synthetic_stats}
\end{table*}

\section{Appendix: Prompt for Noise Data Curation} \label{appendix_curation}

\begin{figure*}[ht]
\centering
\fbox{%
  \begin{minipage}{16cm}
    \textbf{SYSTEM PROMPT} \\
    You are a translation quality analyst. Your job is to determine whether a machine translation output contains ONLY the translation, or whether it also contains extra content that should NOT be part of a clean translation. \\
    
    You will be given: \\
    - source: the original text \\
    - src\_lang / tgt\_lang: language codes \\
    - reference: a clean reference translation \\
    - translation: the LLM-generated translation to judge \\
    
    A ``noisy'' translation contains one or more of these artifacts: \\
    
    1. language\_prefix: A language name label before the translation, e.g. ``Chinese: \foreignlanguage{chinese}{你好}'' \\
    2. verbose\_preamble: An introductory sentence like ``Here is the translation...'' or ``Sure! Here's...'' \\
    3. translation\_prefix: A ``Translation:'' or ``Translated:'' label \\
    4. explanation: Word-by-word breakdown, pinyin/romanization, grammar notes, or extended commentary after the translation (usually separated by newlines) \\
    5. cultural\_note: Usually a parenthetical ``(Note: ...)'' explaining cultural context, idioms, or translation choices \\
    6. bilingual\_output: Both source and target languages appear with labels, or the source text is substantially repeated \\
    7. alternatives: Multiple numbered translation options \\
    8. code\_block: Translation wrapped in markdown code fences (\textasciigrave \textasciigrave \textasciigrave) \\
    9. special\_formatting: Double brackets [[...]], double braces \{\{...\}\}, or XML-like tags \\
    10. extra\_punctuation: Excessive repeated punctuation like ``!!!'' or ``???'' that isn't in the source \\
    11. wrong\_language: The translation is in the wrong language entirely (not the target language) \\
    12. off\_topic: The output is completely unrelated to translation (e.g., code, random text, instructions) \\
    
    A ``clean'' translation contains ONLY the translated text in the target language, possibly with minor differences from the reference (which is fine — different valid translations exist). \\
    
    IMPORTANT: Minor differences in word choice, sentence structure, or style between the translation and the reference do NOT make it noisy. Only extra non-translation content counts. \\
    
    Respond with ONLY valid JSON in this exact format:\\
    \{``is\_noisy'': true or false, \\
      ``confidence'': ``high'' or ``medium'' or ``low'', \\
      ``noise\_patterns'': [``pattern1'', ``pattern2''] or [], \\
      ``primary\_pattern'': ``the most prominent pattern'' or null, \\
      ``reasoning'': ``brief explanation in one sentence'' \}

    \textbf{USER PROMPT} \\
    Source (\{src\_lang\} → \{tgt\_lang\}): \\
    \{source\} \\
    Reference translation: \\
    \{reference\} \\
    LLM translation to judge: \\
    \{translation\} \\
    Is this translation noisy? Respond with JSON only.
  \end{minipage}
  }
  \caption{Prompt for curating noise data.}
\label{noise_curation}
\end{figure*}

\section{Appendix: Prompt for Clean Translation Annotation} \label{appendix_anno}

\begin{figure*}[ht]
\centering
\fbox{%
  \begin{minipage}{16cm}
    \textbf{SYSTEM PROMPT} \\
    You are a translation quality expert. Your task is to extract the clean translation from a noisy machine translation output. \\
    
    The noisy output may contain artifacts such as: \\
    - Language prefixes or labels (e.g. ``Chinese: ...'') \\
    - Verbose preambles (e.g. ``Here is the translation...'') \\
    - Explanations, grammar notes, or word-by-word breakdowns \\
    - Cultural notes or parenthetical comments \\
    - Multiple alternative translations \\
    - Bilingual output with both source and target text \\
    - Code blocks, special formatting, or extra punctuation \\
    - Wrong language or off-topic content \\
    
    You will be given the source text, language pair, a reference translation, the noisy LLM output, and the identified noise patterns. Use all of this context to extract ONLY the clean translation in the target language. \\
    
    If the output contains multiple translation alternatives, extract the best one. If the output is entirely off-topic or in the wrong language, return an empty string. \\
    
    Respond with ONLY valid JSON in this exact format: \\
    \{``extracted\_translation'': ``the clean translation text only''\} \\
    \textbf{USER PROMPT} \\
    Source (\{src\_lang\} → \{tgt\_lang\}): \\
    \{source\} \\
    Reference translation: \\
    \{reference\} \\
    Identified noise patterns: \{noise\_patterns\} \\
    Noisy LLM output to clean: \\
    \{translation\} \\
    
    Extract the clean translation. Respond with JSON only.
  \end{minipage}
  }
  \caption{Prompt for annotating clean translation.}
\label{clean_anno}
\end{figure*}

\section{Appendix: Details for Running Noise Extractors} \label{appendix_ext_details}

We used vLLM for running LLM-based extraction methods with temperature set as $0$ and top\_p $1.0$ on 4 NVIDIA GH200 GPUs. For the span-based extraction method, we ran COMET-KIWI via the COMET repository \citep{COMET2025} on one NVIDIA A100 40BG GPU. On average, an LLM takes approximately 25 minutes to process all (9900) instances, whereas the span-based method costs about 70 minutes.

\section{Appendix: Prompt for LLM-based Extraction} \label{appendix_extractor}

\begin{figure*}[ht]
\centering
\fbox{%
  \begin{minipage}{16cm}
    \textbf{USER PROMPT} \\
    Analyze this machine translation output. The source language is \{src\_lang\} and the target language is \{tgt\_lang\}. \\
    1. Extract ONLY the clean translation (no explanations, labels, or formatting). \\
    2. Identify the noise pattern if the output contains noise. \\

    Return a JSON object with these fields: \\
    - ``extracted\_translation'': the clean translation text only \\
    - ``noise\_pattern'': one of ``none'', ``language\_prefix'', ``verbose\_preamble'', ''translation\_prefix'', ``explanation'', ``cultural\_note'', ``bilingual\_output'', ``alternatives'', ``code\_block'', ``special\_formatting'', ``extra\_punctuation'', ``wrong\_language'', ``off\_topic'' \\
    
    Machine translation output: \{translation\} \\
    JSON:
  \end{minipage}
  }
  \caption{Prompt for LLM-based Extraction.}
\label{prompt_llm_ext}
\end{figure*}

\end{document}